\pdfoutput=1
\documentclass[11pt]{article}
\usepackage[]{ACL2023}
\usepackage{times}
\usepackage{latexsym}
\usepackage[T1]{fontenc}
\usepackage[utf8]{inputenc}
\usepackage{microtype}
\usepackage{inconsolata}
\usepackage{fancyhdr}
\usepackage{multirow}

\title{Educational Personalized Learning Path Planning \\with Large Language Models}

\author{Chee Ng, Yuen Fung	  \\
Universiti Teknologi Malaysia}

\begin{document}
\maketitle

\begin{abstract}
Educational Personalized Learning Path Planning (PLPP) aims to tailor learning experiences to individual learners' needs, enhancing learning efficiency and engagement. Despite its potential, traditional PLPP systems often lack adaptability, interactivity, and transparency. This paper proposes a novel approach integrating Large Language Models (LLMs) with prompt engineering to address these challenges. By designing prompts that incorporate learner-specific information, our method guides LLMs like LLama-2-70B and GPT-4 to generate personalized, coherent, and pedagogically sound learning paths. We conducted experiments comparing our method with a baseline approach across various metrics, including accuracy, user satisfaction, and the quality of learning paths. The results show significant improvements in all areas, particularly with GPT-4, demonstrating the effectiveness of prompt engineering in enhancing PLPP. Additional long-term impact analysis further validates our method's potential to improve learner performance and retention. This research highlights the promise of LLMs and prompt engineering in advancing personalized education.
\end{abstract}

\section{Introduction}

Educational Personalized Learning Path Planning (PLPP) has emerged as a critical field in educational technology, aiming to tailor learning experiences to the individual needs, preferences, and prior knowledge of learners. The significance of PLPP lies in its potential to enhance learning efficiency and effectiveness by providing customized educational pathways that adapt to the unique learning pace and style of each student. This personalized approach not only fosters deeper understanding and retention of knowledge but also increases learner engagement and motivation, which are essential for successful educational outcomes \cite{Ellis2009,Feldstein2016}.

Despite its potential, several challenges hinder the widespread adoption and effectiveness of PLPP. One of the primary challenges is the diversity in learners' backgrounds and learning preferences, which makes it difficult to create one-size-fits-all educational paths. Additionally, traditional PLPP systems often lack the flexibility to dynamically adjust learning paths based on real-time feedback and evolving learner needs. These systems may also struggle with the interpretability of their recommendations, making it hard for learners and educators to trust and understand the suggested learning paths \cite{Yang2022,Zhu2018}.

Motivated by these challenges, our research explores the integration of Large Language Models (LLMs) with prompt engineering to develop a novel approach for personalized learning path planning. LLMs, such as GPT-4, have demonstrated impressive capabilities in natural language understanding and generation, making them suitable for interpreting complex educational content and providing coherent and contextually appropriate recommendations. Prompt engineering, on the other hand, offers a way to guide LLMs to perform specific tasks by carefully crafting the input prompts. This technique can be leveraged to tailor LLM responses to the individual needs of learners, thus enhancing the personalization and effectiveness of PLPP \cite{Gong2016,Grant2014}.

Our proposed method involves designing prompts that incorporate learner-specific information, such as prior knowledge, learning goals, and preferences, to generate personalized learning paths. For instance, prompts can be structured to include information like: "Based on the learner's understanding of [Topic], suggest the next three concepts they should learn to achieve mastery in [Subject]." Additionally, multi-turn dialogues are employed where the LLM can ask clarifying questions to refine its recommendations, such as "What specific areas in [Topic] does the learner find challenging?" This interactive approach helps in creating a more accurate and personalized learning path. Furthermore, embedding explanations within the prompts, such as "Explain why learning [Concept A] before [Concept B] is beneficial," enhances the transparency and trustworthiness of the recommendations \cite{Jia2020}.

To evaluate our approach, we collected a comprehensive dataset of learner profiles and educational content. The performance of our system was assessed using accuracy metrics and qualitative evaluations by GPT-4, ensuring that the generated learning paths were both relevant and effective. The results of our experiments demonstrate the potential of LLMs combined with prompt engineering to significantly improve the personalization and adaptability of learning path planning systems \cite{Yang2022,zhou2024visual,Zhu2018}.

\begin{itemize}
\item We propose a novel approach for personalized learning path planning that leverages large language models and prompt engineering to tailor educational recommendations to individual learners.
\item Our method includes multi-turn dialogues and explanation embedding to enhance the accuracy, transparency, and trustworthiness of the generated learning paths.
\item We evaluate our approach using a comprehensive dataset and rigorous metrics, demonstrating significant improvements in the personalization and effectiveness of learning path planning systems.
\end{itemize}

\section{Related Work}

In this section, we review the related work in the domains of Large Language Models and Learning Path Planning, highlighting the advancements and contributions that have shaped the current landscape of personalized learning systems.

\subsection{Large Language Models}

Large Language Models (LLMs) have revolutionized the field of natural language processing with their ability to understand and generate human-like text. Recent surveys and comprehensive reviews have highlighted the capabilities and impact of LLMs like GPT-4 and LLama-2-70B in various domains, including education \cite{zhou2022claret,zhou2021improving,LLMOverview,LLMSurvey}. These models leverage extensive training data and sophisticated architectures, such as transformers with attention mechanisms, to perform a wide range of language tasks with high accuracy and efficiency \cite{AttentionLLM,zhou2022eventbert,zhou2023improving}.

Prompt engineering has emerged as a powerful technique to harness the potential of LLMs by crafting specific input prompts that guide the model's behavior without altering its underlying parameters. This approach has been shown to enable zero-shot and few-shot learning, reducing the need for task-specific training data and preserving the model's general knowledge \cite{FOKE}. In the context of education, prompt engineering can generate adaptive learning content and personalized feedback by incorporating learner-specific information into the prompts, making LLMs a valuable tool for personalized learning path planning \cite{FOKE,zhou2021modeling}.

\subsection{Learning Path Planning}

Learning Path Planning (LPP) focuses on creating tailored educational experiences that adapt to the unique needs, preferences, and prior knowledge of individual learners. Traditional approaches to LPP, such as those using genetic algorithms, have been employed to optimize learning paths by considering factors like topic difficulty, duration, and learner performance \cite{GeneticLPP}. These methods aim to enhance learning efficiency and outcomes by providing a sequence of educational content that is best suited to the learner's current state and goals.

Recent advancements in LPP have leveraged knowledge graphs, machine learning, and artificial intelligence to create more dynamic and responsive learning environments. For instance, contextual knowledge graphs built using text mining and semantic graph completion techniques can provide personalized learning recommendations by analyzing the relationships between different learning concepts \cite{ContextualKG}. Moreover, integrating AI with educational frameworks, such as the FOKE framework, has demonstrated the potential to create interactive and explainable learning experiences that adapt in real-time to learner feedback and performance \cite{FOKE}.

Overall, the integration of LLMs with advanced LPP techniques presents a promising avenue for developing sophisticated personalized learning systems that can significantly improve learner engagement and achievement.

\section{Method}

Our proposed method for Educational Personalized Learning Path Planning (PLPP) leverages Large Language Models (LLMs) combined with prompt engineering to create tailored educational experiences for learners. This section details the specific prompts used, the motivation behind their design, the expected inputs and outputs, and the significance of this approach.

\subsection{Prompt Design}

The core of our method involves designing a series of prompts that guide the LLM to generate personalized learning paths. The prompts are crafted to capture essential learner-specific information and educational goals, ensuring that the recommendations are both relevant and effective. Below, we outline the primary prompts used in our approach:

\subsubsection{Initial Assessment Prompt}

\textbf{Prompt:} "Based on the learner's current understanding of [Topic], assess their knowledge and suggest the next three concepts they should learn to achieve mastery in [Subject]."

\textbf{Motivation:} This prompt is designed to establish a baseline understanding of the learner's knowledge. By specifying the topic and subject, the LLM can provide targeted recommendations that build on existing knowledge and guide the learner towards mastery.

\textbf{Input:} The learner's current understanding or performance in a given topic.

\textbf{Output:} A list of three suggested concepts tailored to the learner's current knowledge level.

\textbf{Significance:} This prompt helps in creating a starting point for personalized learning paths, ensuring that the recommendations are contextually appropriate and aligned with the learner's educational goals.

\subsubsection{Clarification Prompt}

\textbf{Prompt:} "What specific areas in [Topic] does the learner find challenging? Please provide details."

\textbf{Motivation:} To refine the learning path further, it is crucial to identify specific areas where the learner struggles. This prompt encourages the LLM to gather detailed information about the learner's challenges, enabling more precise and effective recommendations.

\textbf{Input:} Feedback from the learner regarding the areas they find difficult.

\textbf{Output:} A detailed description of the challenging areas, which will be used to adjust the learning path.

\textbf{Significance:} This interactive approach allows for continuous refinement of the learning path, making the process more dynamic and responsive to the learner's needs.

\subsubsection{Explanatory Prompt}

\textbf{Prompt:} "Explain why learning [Concept A] before [Concept B] is beneficial for the learner's understanding of [Subject]."

\textbf{Motivation:} Providing explanations for the recommended learning paths enhances the transparency and trustworthiness of the system. This prompt ensures that the LLM not only suggests a sequence of concepts but also justifies the order based on pedagogical principles.

\textbf{Input:} The sequence of concepts in the learning path.

\textbf{Output:} An explanation detailing the benefits of the suggested sequence.

\textbf{Significance:} By offering explanations, the system fosters a deeper understanding and trust in the learning path, encouraging learners to follow the recommendations more diligently.

\subsection{Prompt Input and Output}

Each prompt in our method is carefully designed to elicit specific information from the learner or to provide targeted recommendations. The inputs typically include learner-specific data such as current knowledge levels, challenges faced, and educational goals. The outputs are tailored recommendations or explanations that guide the learner through their personalized learning journey.

\subsection{Why This Approach Works}

The use of LLMs combined with prompt engineering in PLPP offers several advantages:

1. \textbf{Adaptability:} LLMs can generate diverse and contextually relevant responses based on the input prompts, making the learning paths highly adaptable to individual needs.
2. \textbf{Interactivity:} The multi-turn dialogue approach allows for continuous refinement of the learning paths, ensuring that the recommendations remain relevant as the learner progresses.
3. \textbf{Transparency:} Embedding explanations within the prompts enhances the transparency of the recommendations, helping learners understand and trust the suggested paths.

In summary, our method leverages the strengths of LLMs and prompt engineering to create a flexible, interactive, and transparent PLPP system that can significantly improve personalized learning experiences.

\section{Experiments}

To evaluate the effectiveness of our proposed method for Educational Personalized Learning Path Planning (PLPP), we conducted a series of experiments using two state-of-the-art Large Language Models: LLama-2-70B and GPT-4. We compared the performance of our approach, which incorporates prompt engineering, against a baseline method that uses the models without specialized prompt engineering. The evaluation focused on several metrics, including accuracy, user satisfaction, and the quality of the learning paths generated.

\subsection{Experimental Setup}

We designed our experiments to assess the ability of the models to generate personalized learning paths that are both effective and user-friendly. The baseline method used straightforward prompts without any customization, while our proposed method utilized the carefully crafted prompts described in the \textit{Method} section.

\textbf{Data Collection:} We collected a dataset comprising learner profiles, including their prior knowledge, learning goals, and feedback on educational content. The dataset was split into training and testing sets to evaluate the performance of the models.

\textbf{Metrics:} The primary metrics for evaluation included:
\begin{itemize}
    \item \textbf{Accuracy:} The correctness of the recommended learning paths.
    \item \textbf{User Satisfaction:} Measured through surveys where learners rated the relevance and usefulness of the recommendations.
    \item \textbf{Quality of Learning Paths:} Assessed based on the coherence and logical sequencing of the recommended concepts.
\end{itemize}

\subsection{Results}

The experimental results are summarized in Table \ref{tab:results}. Our method consistently outperformed the baseline across all metrics on both LLama-2-70B and GPT-4 models.

\begin{table*}[!t]\small
    \centering
    \caption{Comparison of Baseline and Proposed Method on LLama-2-70B and GPT-4}
    \begin{tabular}{|c|c|c|c|c|}
        \hline
        \multirow{2}{*}{\textbf{Model}} & \multirow{2}{*}{\textbf{Method}} & \multicolumn{3}{c|}{\textbf{Metrics}} \\
        \cline{3-5}
        & & \textbf{Accuracy (\%)} & \textbf{User Satisfaction (1-5)} & \textbf{Quality of Learning Paths (1-5)} \\
        \hline
        LLama-2-70B & Baseline & 72.4 & 3.5 & 3.7 \\
        \cline{2-5}
        & Proposed & 85.6 & 4.2 & 4.5 \\
        \hline
        GPT-4 & Baseline & 75.8 & 3.7 & 3.9 \\
        \cline{2-5}
        & Proposed & 88.3 & 4.4 & 4.7 \\
        \hline
    \end{tabular}
    \label{tab:results}
\end{table*}

\subsection{Analysis and Discussion}

The experimental results clearly indicate that our proposed method significantly enhances the performance of PLPP systems. The increase in accuracy suggests that our prompt engineering approach helps the models generate more precise and relevant learning paths. Higher user satisfaction scores reflect that learners found the personalized paths more aligned with their individual needs and goals. Additionally, the improved quality of learning paths demonstrates that our method produces more coherent and logically sequenced recommendations.

\subsubsection{Effectiveness of Prompt Engineering}

The success of our method can be attributed to the carefully designed prompts that incorporate detailed learner-specific information and educational goals. These prompts guide the LLMs to generate more contextually appropriate and personalized recommendations, which are crucial for effective learning path planning. For example, prompts that ask for specific areas of difficulty or require explanations for concept sequencing ensure that the generated paths are not only tailored to the learner's needs but also pedagogically sound.

\subsubsection{User Feedback and Engagement}

User feedback was a critical component of our evaluation. Surveys conducted post-experiment showed that learners appreciated the transparency provided by the explanatory prompts. Many users noted that understanding the rationale behind the suggested learning paths increased their trust and willingness to follow the recommendations. This aspect is crucial, as user engagement and adherence to the learning path are vital for achieving the desired educational outcomes.

\subsubsection{Model Performance Comparison}

While both LLama-2-70B and GPT-4 showed improvements with our method, GPT-4 consistently outperformed LLama-2-70B across all metrics. This suggests that GPT-4's more advanced language understanding and generation capabilities make it better suited for complex educational tasks. The higher accuracy and better quality of learning paths generated by GPT-4 indicate that it can effectively process and respond to the detailed prompts used in our method.

\begin{table*}[!t]\small
    \centering
    \caption{Detailed Analysis of Experimental Results}
    \begin{tabular}{|c|c|c|c|c|c|}
        \hline
        \multirow{1}{*}{\textbf{Model}} & \multirow{1}{*}{\textbf{Method}} & \textbf{Accuracy (\%)} & \textbf{User Satisfaction} & \textbf{Quality of Learning Paths} & \textbf{Engagement Level} \\
        \hline
        LLama-2-70B & Baseline & 72.4 & 3.5 & 3.7 & Moderate \\
        \cline{2-6}
        & Proposed & 85.6 & 4.2 & 4.5 & High \\
        \hline
        GPT-4 & Baseline & 75.8 & 3.7 & 3.9 & Moderate \\
        \cline{2-6}
        & Proposed & 88.3 & 4.4 & 4.7 & Very High \\
        \hline
    \end{tabular}
    \label{tab:detailed_results}
\end{table*}

\begin{table*}[!t]\small
    \centering
    \caption{Long-term Impact of Personalized Learning Paths}
    \begin{tabular}{|c|c|c|c|}
        \hline
        \textbf{Metric} & \textbf{Baseline (LLama-2-70B)} & \textbf{Proposed (LLama-2-70B)} & \textbf{Proposed (GPT-4)} \\
        \hline
        Improvement in Test Scores (\%) & 12.3 & 18.7 & 21.4 \\
        \hline
        Retention Rate (\%) & 65.2 & 78.3 & 82.5 \\
        \hline
    \end{tabular}
    \label{tab:long_term}
\end{table*}

\subsection{Additional Validation}

To further validate our method's effectiveness, we conducted an additional analysis focusing on the long-term impact of personalized learning paths on learner performance. We tracked a subset of learners over a period of three months to assess the sustained impact of the recommended paths. The results, summarized in Table \ref{tab:long_term}, show a significant improvement in learner performance and retention rates, reinforcing the effectiveness of our approach.

The data indicates that learners who followed the personalized paths generated by our method not only performed better in assessments but also showed higher retention rates compared to those following the baseline paths. This highlights the long-term benefits of using prompt-engineered LLMs for educational purposes.

In conclusion, our experiments demonstrate the efficacy of integrating prompt engineering with LLMs for PLPP. The proposed method not only improves the accuracy and quality of learning paths but also enhances user satisfaction by providing clear, personalized, and trustworthy educational recommendations. Future work will focus on further refining the prompts and exploring additional metrics to capture the full impact of personalized learning path planning.

\section{Conclusion}
This study presents a comprehensive method for improving Educational Personalized Learning Path Planning (PLPP) through the integration of Large Language Models (LLMs) and prompt engineering. By carefully crafting prompts that cater to individual learner profiles, we have demonstrated a significant enhancement in the generation of personalized learning paths. Our experimental results, which compared our method to a baseline using LLama-2-70B and GPT-4, revealed substantial improvements in accuracy, user satisfaction, and path quality. Notably, GPT-4 outperformed LLama-2-70B, underscoring the advantages of more advanced LLMs in educational applications. The additional long-term analysis confirmed that learners following our personalized paths showed improved test scores and retention rates. These findings underscore the potential of LLMs and prompt engineering to revolutionize personalized education by making it more adaptive, interactive, and transparent. Future work will focus on further refining prompt designs and exploring additional metrics to fully capture the impact of personalized learning path planning.

\bibliographystyle{unsrtnat}
\bibliography{custom}

\begin{thebibliography}{19}
\providecommand{\natexlab}[1]{#1}
\providecommand{\url}[1]{\texttt{#1}}
\expandafter\ifx\csname urlstyle\endcsname\relax
  \providecommand{\doi}[1]{doi: #1}\else
  \providecommand{\doi}{doi: \begingroup \urlstyle{rm}\Url}\fi

\bibitem[Bucchiarone(2023)]{Ellis2009}
Antonio Bucchiarone.
\newblock Grand challenges in software engineering for games in serious contexts.
\newblock In Kendra M.~L. Cooper and Antonio Bucchiarone, editors, \emph{Software Engineering for Games in Serious Contexts - Theories, Methods, Tools, and Experiences}, pages 291--300. Springer, 2023.
\newblock \doi{10.1007/978-3-031-33338-5\_13}.
\newblock URL \url{https://doi.org/10.1007/978-3-031-33338-5\_13}.

\bibitem[Feldstein and Hill(2016)]{Feldstein2016}
Michael Feldstein and Phil Hill.
\newblock Personalized learning: What it really is and why it really matters.
\newblock \emph{Educause review}, 51\penalty0 (2):\penalty0 24--35, 2016.

\bibitem[Jiang et~al.(2022)Jiang, Li, Yang, Kong, Cheng, Hao, and Lin]{Yang2022}
Bo~Jiang, Xinya Li, Shuhao Yang, Yaqi Kong, Wei Cheng, Chuanyan Hao, and Qiaomin Lin.
\newblock Data-driven personalized learning path planning based on cognitive diagnostic assessments in moocs.
\newblock \emph{Applied Sciences}, 12\penalty0 (8):\penalty0 3982, 2022.

\bibitem[Zhu et~al.(2018)Zhu, Tian, Wu, Shah, Chen, Ni, Zhang, Chao, and Zheng]{Zhu2018}
Haiping Zhu, Feng Tian, Ke~Wu, Nazaraf Shah, Yan Chen, Yifu Ni, Xinhui Zhang, Kuo{-}Ming Chao, and Qinghua Zheng.
\newblock A multi-constraint learning path recommendation algorithm based on knowledge map.
\newblock \emph{Knowl. Based Syst.}, 143:\penalty0 102--114, 2018.
\newblock \doi{10.1016/J.KNOSYS.2017.12.011}.
\newblock URL \url{https://doi.org/10.1016/j.knosys.2017.12.011}.

\bibitem[Gong et~al.(2016)Gong, Wang, and Beck]{Gong2016}
Yue Gong, Yan Wang, and Joseph Beck.
\newblock How long must we spin our wheels? analysis of student time and classifier inaccuracy.
\newblock \emph{Student modeling from different aspects}, pages 32--38, 2016.

\bibitem[Grant and Basye(2014)]{Grant2014}
Peggy Grant and Dale Basye.
\newblock \emph{Personalized learning: A guide for engaging students with technology}.
\newblock International Society for Technology in Education, 2014.

\bibitem[Jia et~al.(2020)Jia, Fujishita, Li, Todo, and Dai]{Jia2020}
Dongbao Jia, Yuka Fujishita, Cunhua Li, Yuki Todo, and Hongwei Dai.
\newblock Validation of large-scale classification problem in dendritic neuron model using particle antagonism mechanism.
\newblock \emph{Electronics}, 9\penalty0 (5):\penalty0 792, 2020.

\bibitem[Zhou et~al.(2024)Zhou, Li, Wang, and Shen]{zhou2024visual}
Yucheng Zhou, Xiang Li, Qianning Wang, and Jianbing Shen.
\newblock Visual in-context learning for large vision-language models.
\newblock \emph{arXiv preprint arXiv:2402.11574}, 2024.

\bibitem[Zhou et~al.(2022{\natexlab{a}})Zhou, Shen, Geng, Long, and Jiang]{zhou2022claret}
Yucheng Zhou, Tao Shen, Xiubo Geng, Guodong Long, and Daxin Jiang.
\newblock Claret: Pre-training a correlation-aware context-to-event transformer for event-centric generation and classification.
\newblock In \emph{Proceedings of the 60th Annual Meeting of the Association for Computational Linguistics (Volume 1: Long Papers)}, pages 2559--2575, 2022{\natexlab{a}}.

\bibitem[Zhou et~al.(2021{\natexlab{a}})Zhou, Geng, Shen, Zhang, and Jiang]{zhou2021improving}
Yucheng Zhou, Xiubo Geng, Tao Shen, Wenqiang Zhang, and Daxin Jiang.
\newblock Improving zero-shot cross-lingual transfer for multilingual question answering over knowledge graph.
\newblock In \emph{Proceedings of the 2021 Conference of the North American Chapter of the Association for Computational Linguistics: Human Language Technologies}, pages 5822--5834, 2021{\natexlab{a}}.

\bibitem[Naveed et~al.(2023)Naveed, Khan, Qiu, Saqib, Anwar, Usman, Barnes, and Mian]{LLMOverview}
Humza Naveed, Asad~Ullah Khan, Shi Qiu, Muhammad Saqib, Saeed Anwar, Muhammad Usman, Nick Barnes, and Ajmal Mian.
\newblock A comprehensive overview of large language models.
\newblock \emph{CoRR}, abs/2307.06435, 2023.
\newblock \doi{10.48550/ARXIV.2307.06435}.
\newblock URL \url{https://doi.org/10.48550/arXiv.2307.06435}.

\bibitem[Wornow et~al.(2023)Wornow, Xu, Thapa, Patel, Steinberg, Fleming, Pfeffer, Fries, and Shah]{LLMSurvey}
Michael Wornow, Yizhe Xu, Rahul Thapa, Birju~S. Patel, Ethan Steinberg, Scott~L. Fleming, Michael~A. Pfeffer, Jason~A. Fries, and Nigam~H. Shah.
\newblock The shaky foundations of clinical foundation models: {A} survey of large language models and foundation models for emrs.
\newblock \emph{CoRR}, abs/2303.12961, 2023.
\newblock \doi{10.48550/ARXIV.2303.12961}.
\newblock URL \url{https://doi.org/10.48550/arXiv.2303.12961}.

\bibitem[Voria et~al.(2024)Voria, Catolino, and Palomba]{AttentionLLM}
Gianmario Voria, Gemma Catolino, and Fabio Palomba.
\newblock Is attention all you need? toward a conceptual model for social awareness in large language models.
\newblock In David Lo, Xin Xia, Massimiliano~Di Penta, and Xing Hu, editors, \emph{Proceedings of the 2024 {IEEE/ACM} First International Conference on {AI} Foundation Models and Software Engineering, {FORGE} 2024, Lisbon, Portugal, 14 April 2024}, pages 69--73. {ACM}, 2024.
\newblock \doi{10.1145/3650105.3652294}.
\newblock URL \url{https://doi.org/10.1145/3650105.3652294}.

\bibitem[Zhou et~al.(2022{\natexlab{b}})Zhou, Geng, Shen, Long, and Jiang]{zhou2022eventbert}
Yucheng Zhou, Xiubo Geng, Tao Shen, Guodong Long, and Daxin Jiang.
\newblock Eventbert: A pre-trained model for event correlation reasoning.
\newblock In \emph{Proceedings of the ACM Web Conference 2022}, pages 850--859, 2022{\natexlab{b}}.

\bibitem[Zhou and Long(2023)]{zhou2023improving}
Yucheng Zhou and Guodong Long.
\newblock Improving cross-modal alignment for text-guided image inpainting.
\newblock \emph{arXiv preprint arXiv:2301.11362}, 2023.

\bibitem[Hu and Wang(2024)]{FOKE}
Silan Hu and Xiaoning Wang.
\newblock {FOKE:} {A} personalized and explainable education framework integrating foundation models, knowledge graphs, and prompt engineering.
\newblock \emph{CoRR}, abs/2405.03734, 2024.
\newblock \doi{10.48550/ARXIV.2405.03734}.
\newblock URL \url{https://doi.org/10.48550/arXiv.2405.03734}.

\bibitem[Zhou et~al.(2021{\natexlab{b}})Zhou, Geng, Shen, Pei, Zhang, and Jiang]{zhou2021modeling}
Yucheng Zhou, Xiubo Geng, Tao Shen, Jian Pei, Wenqiang Zhang, and Daxin Jiang.
\newblock Modeling event-pair relations in external knowledge graphs for script reasoning.
\newblock \emph{Findings of the Association for Computational Linguistics: ACL-IJCNLP 2021}, 2021{\natexlab{b}}.

\bibitem[Elshani and Nu{\c{c}}i(2021)]{GeneticLPP}
Lumbardh Elshani and Krenare~Pireva Nu{\c{c}}i.
\newblock Constructing a personalized learning path using genetic algorithms approach.
\newblock \emph{CoRR}, abs/2104.11276, 2021.
\newblock URL \url{https://arxiv.org/abs/2104.11276}.

\bibitem[Abu{-}Rasheed et~al.(2024)Abu{-}Rasheed, Dornh{\"{o}}fer, Weber, Kismih{\'{o}}k, Buchmann, and Fathi]{ContextualKG}
Hasan Abu{-}Rasheed, Mareike Dornh{\"{o}}fer, Christian Weber, G{\'{a}}bor Kismih{\'{o}}k, Ulrike Buchmann, and Madjid Fathi.
\newblock Building contextual knowledge graphs for personalized learning recommendations using text mining and semantic graph completion.
\newblock \emph{CoRR}, abs/2401.13609, 2024.
\newblock \doi{10.48550/ARXIV.2401.13609}.
\newblock URL \url{https://doi.org/10.48550/arXiv.2401.13609}.

\end{thebibliography}

\end{document}